\documentclass[10pt,conference,a4paper,conference]{IEEEtran}

\usepackage{cite}
\usepackage{amsmath,amssymb,amsfonts}
\usepackage[colorlinks=true, pagebackref]{hyperref}  %pagebackref hidelinks hyperref still needs to be put at the end!
\usepackage[square,sort&compress,comma,numbers]{natbib}% space in references ch (sort&compress for [1-4])
\usepackage{graphicx}
\usepackage{textcomp}
\usepackage{xcolor}
\usepackage{color,soul}
\usepackage[normalem]{ulem}
\usepackage{authblk}
\usepackage{diagbox}
\usepackage{pdfcomment}
\usepackage{todonotes}
\usepackage{multirow}
\usepackage{booktabs}
\usepackage{pifont}% http://ctan.org/pkg/pifont

\def\BibTeX{{\rm B\kern-.05em{\sc i\kern-.025em b}\kern-.08em
		T\kern-.1667em\lower.7ex\hbox{E}\kern-.125emX}}

\newcommand\redst{\bgroup\markoverwith{\textcolor{red}{\rule[0.5ex]{2pt}{1pt}}}\ULon}

\begin{document}

	\title{Competitive Simplicity for Multi-Task Learning for Real-Time Foggy Scene Understanding via \\Domain Adaptation} % Replace with your title

\author[1, 2]{Naif Alshammari}
\author[1, 3]{Samet Ak\c{c}ay}
\author[1, 4]{Toby P. Breckon\vspace{-.15cm}}

\affil[1]{ 
	Department of Computer Science, Durham University, Durham, UK
}%\vspace{1.5ex}}

\affil[2]{
	Department of Natural and Applied Science, Majma\textquotesingle ah University, Majma\textquotesingle ah, KSA}
\affil[3]{COSMONiO, Durham, UK}
\affil[4]{
	Department of Engineering, Durham University, Durham, UK \authorcr 
	{\tt\small
		\{\href{mailto:naif.alshammari@durham.ac.uk}{naif.alshammari},
		\href{mailto:samet.akcay@durham.ac.uk}{samet.akcay},
		\href{mailto:toby.breckon@durham.ac.uk}{toby.breckon}\}@durham.ac.uk
	}	
}

\maketitle

\begin{abstract} %Adapting between two domains (\textit{normal} and \textit{foggy}),
Automotive scene understanding under adverse weather conditions raises a realistic and challenging problem attributable to poor outdoor scene visibility (e.g. foggy weather). However, because most contemporary scene understanding approaches are applied under ideal-weather conditions, such approaches may not provide genuinely optimal performance when compared to established \textit{a priori} insights on extreme-weather understanding. In this paper, we propose a complex but competitive multi-task learning approach capable of performing in real-time semantic scene understanding and monocular depth estimation under \textit{foggy} weather conditions by leveraging both recent advances in adversarial training and domain adaptation. As an end-to-end pipeline, our model provides a novel solution to surpass degraded visibility in \textit{foggy} weather conditions by 
transferring scenes from \textit{foggy} to \textit{normal} using a GAN-based model. For optimal performance in semantic segmentation, our model generates depth to be used as complementary source information with RGB in the segmentation network. We provide a robust method for \textit{foggy} scene understanding by training two models (\textit{normal} and \textit{foggy}) simultaneously with shared weights (each model is trained on each weather condition independently). Our model incorporates RGB colour, depth, and luminance images via distinct encoders with dense connectivity and features fusing, and leverages skip connections to produce consistent depth and segmentation predictions. Using this architectural formulation with light computational complexity at inference time, we are able to achieve comparable performance to contemporary approaches at a fraction of the overall model complexity. Evaluation over several foggy weather condition datasets including synthetic and real-world examples illustrates
our approach competitive performance compared to other contemporary state-of-the-art approaches.
\end{abstract}

	\section{Introduction}

\begin{figure}[!tb]
	\centering
	\includegraphics[width=\linewidth]{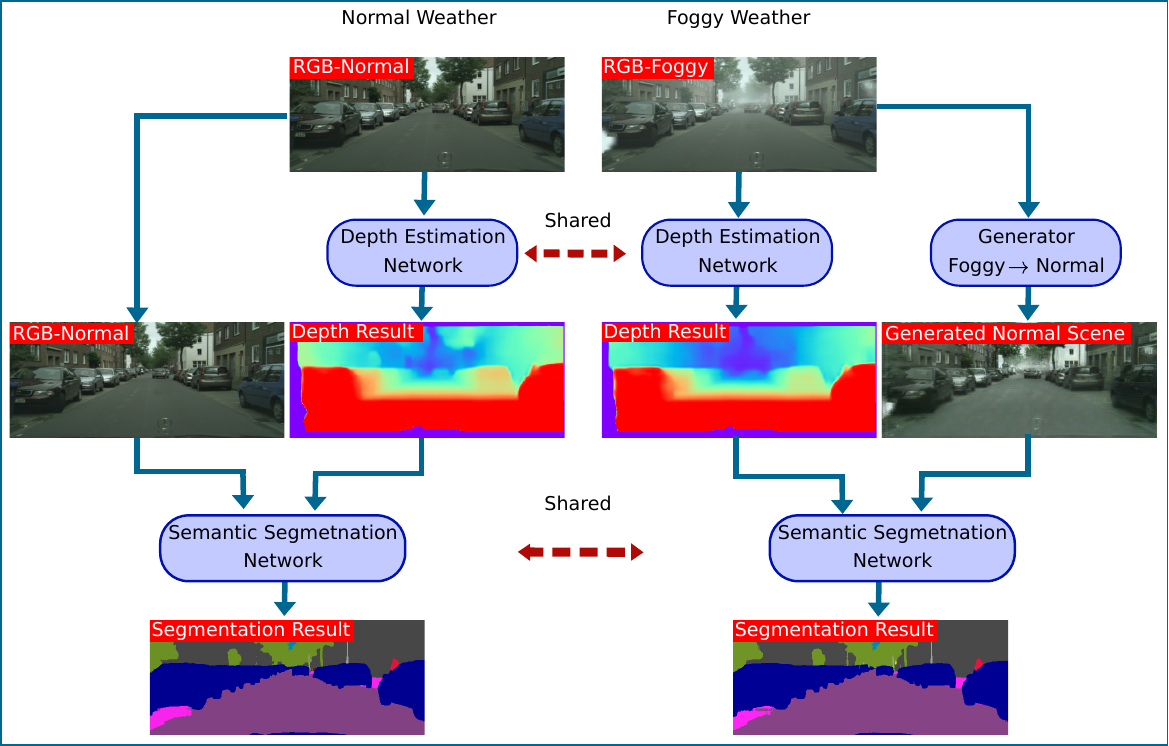}
	\caption{A high-level illustration of our pipeline for semantic segmentation and depth estimation under foggy weather conditions.}
	\vspace{-20px} 
	\label{fig:intro}
\end{figure} 

Semantic segmentation for automotive urban environments is a rapidly developing research topic illustrating successful state-of-the-art scene understanding approaches \cite{segnet17, deeplab, refineNet17, pyramid}. Despite its successes, limited attention has been paid to the issue of automotive scene understanding under extreme weather conditions (\textit{i.e.} foggy weather conditions) \cite{foggy18, foggy19}, and by contrast we see deep learning approaches generally applicable to ideal weather conditions only. This paper proposes a robust solution to this challenge by taking advantage of domain adaptation for transferring knowledge from one domain to another -- in this case, the between the domains of \textit{normal} and \textit{foggy} scene weather conditions.

Previous approaches for reducing adverse weather impact on automotive has seen differing methods proposed for reducing the illumination variance \cite{naif, kim2017pca}, adapting scene understanding methods from day to night \cite{sakaridis2019guided} or synthetic fog \cite{foggy18, foggy19}. Following recent advances in deep learning, scene understanding under such challenging conditions has also been addressed via domain adaptation \cite{wenzel2018modular} where a scene taken in foggy weather conditions is first pre-mapped onto a target domain (clear-weather), which is considered the optimal input for secondary scene understanding approaches. 

In contrast to this pre-transformation approach, here we propose an end-to-end semantic scene understanding and monocular depth estimation framework using a novel multi-task approach specifically targeting the challenge of foggy weather operating conditions directly in the automotive environment. As the main objective of our work, we tackle the issue of semantic segmentation under \textit{foggy} weather conditions in four steps. First, by employing the domain adaptation approach via image style transfer as proposed in \cite{CycleGAN2017} as a method to increase the level of visibility that suffers in \textit{foggy} weather conditions. Second, by taking advantage of complementary depth information generated by a monocular depth estimator, which can be subsequently provided as an additional input to RGB colour into a semantic segmentor. These depth estimations and semantic segmentation components are trained via sub-models on both domains (\textit{normal} and \textit{foggy}) (each domain is trained independently), with shared weights to allow the implicit transfer of semantic and depth knowledge from one domain to another. Finally, our model is adversarially trained on output streams from depth estimation and semantic segmentation to force the multi-task model to produce predictions as close to the target outputs as possible. Figure \ref{fig:intro} shows an illustration of our overall approach, including the steps mentioned above.  

In summary, the main contributions of this paper are as follows:

%(1) transferring scenes from foggy weather conditions to normal
%(i.e. surpassing degraded visibility) using a GAN-based model; 
%
%(2) generating complementary depth information that contributes to improved
%semantic segmentation; 
%
%(3) performing semantic segmentation by taking
%advantage of steps 1 and 2 and knowledge adapted between two domains
%(normal and adverse) trained simultaneously with shared weights (each
%model is trained on each weather condition independently); 
%
%(4) employing adversarial techniques on the output streams of the above-mentioned
%semantic segmentation and depth estimation networks to improve their
%performance.

\vspace{-.05cm}
\begin{itemize}
    \itemsep0em
    
    \vspace{.15cm}
     \item {\itshape \textit{Competitive low-complexity architecture -- }} enables semantic segmentation and depth prediction via  multi-task learning and leveraging domain adaptation to correct images with degraded visibility in \textit{foggy} weather conditions.

     \vspace{.15cm}
     \item {\itshape \textit{Optimal foggy scene understanding -- }} via adapting between two domains (\textit{normal} and \textit{foggy}) trained simultaneously with shared weights (each model is trained on one weather condition independently) and employing adversarial techniques on the output from each model.
     
     \vspace{.15cm}
     \item {\itshape \textit{Competitive performance -- }} outperforms the state-of-the-art foggy scene understanding \cite{foggy19, foggy_pure} on the benchmark datasets \cite{foggy19, foggy18}, hence our model trained on less dataset.
%    
%
%     \vspace{.1cm}
%     \item {\itshape \textit{Competitor performance -- }} running in real-time, performing multi-task and with few number of parameters.}

\end{itemize}

	\section{Related Work}
\label{sec:relatedwork}

We review prior work in three key areas:- semantic segmentation (Section \ref{sec:semseg}), monocular depth estimation (Section \ref{sec:depth}), and domain adaptation (Section \ref{sec:style}).

\subsection{Semantic Segmentation}
\label{sec:semseg}
%\vspace{-.25cm}
Semantic segmentation is an essential task in scene understanding requiring robust per-pixels classification. Prior work has achieved promising results via deep convolutional networks \cite{FCN, unet, refineNet17, pyramid, ldfnet, erfnet}. However, they differ by using different approaches for instance: pooling indices \cite{segnet17}, skip connection \cite{unet}, multi-path refinement \cite{refineNet17}, pyramid pooling \cite{pyramid}, fusing-based \cite{ldfnet} . As the basis  for a number of semantic segmentation architectures, \cite{FCN} leads the recent contributions by adopting \cite{vgg} (an architecture designed for image classification) and subsequently decoding (mapping) low feature representations to pixel-wise output in an end-to-end model. Most prior work on semantic segmentation uses RGB and/or RGB-D data as an input \cite{amir19, segnet17, refineNet17}. As an incorporated technique, other studies have achieved some successes using luminance information \cite{ldfnet, naif}. 

As a key challenge, several different approaches have been proposed to tackle the issue of scene understanding under adverse weather conditions. For instance, the issue of illumination changes is addressed in \cite{kim2017pca, naif} by minimising scene colour variations due to varying scene lighting conditions. Other approaches \cite{foggy_old, foggy18, foggy19} address segmentation under foggy weather conditions using a semi-supervised approach through generating synthetic fog from real-world data and augmenting clear images to their synthetic fog images. By adapting segmentation models from day to night, \cite{sakaridis2019guided} addressed the issue of poor scene visibility. Recently, domain adaptation as a technique (within the context of semantic segmentation) is employed to generate \textit{normal} weather scenes from \textit{adverse} ones \cite{wenzel2018modular} (this can be considered to be a defogging process) using \cite{pix2pix2016, CycleGAN2017} (Section \ref{sec:style}). Subsequently, this generated input is fed into a semantic segmentor to perform pixel-wise segmentation \cite{wenzel2018modular}. 

Another method to achieve improved segmentation \cite{tsai2018learning},  propose a discriminator network using GAN \cite{gan} to encourage a segmentation model, with shared weights between two sub-models trained on different domains (real-world and synthetic images) independently, to produce pixel-wise class labels.

Similarly, our semantic segmentation component is trained via two sub-models (each model on one weather condition independently) using real-world input representing \textit{normal} weather conditions and synthetic \textit{normal} inputs generated from a synthetic foggy dataset using domain adaptation \cite{CycleGAN2017} (discussed in Section \ref{sec:style}). However, inspired by \cite{ldfnet}, we employ the idea of incorporating luminance and depth alongside RGB colour via distinct encoders, utilising both skip connections and dense connectivity as well as fused features to gain better and deeper representation learning which leads to a superior semantic segmentation performance. 
       
\subsection{Monocular Depth Estimation}
\label{sec:depth}
%\vspace{-.25cm}
Although our main objective is semantic segmentation, using monocular depth estimation alongside semantic segmentation via multi-task learning may contribute to achieving better semantic segmentation performance \cite{amir19}. Monocular depth estimation is a technique used to predict depth from a single image. In the literature, monocular depth estimation \cite{garg2016unsupervised, monodepth, Aimr2018, amir19} provides a solution for the shortcomings in depth estimation in terms of either the significant training data requirements or the final output (missing depth) of fundamental strategies \cite{toby_depth}. 

Recent methods addressed monocular depth estimation using both supervised \cite{eigen2014depth, liu2015learning, amir19, Aimr2018} and unsupervised \cite{monodepth, garg2016unsupervised} learning approaches. Employing a Generative Adversarial Network (GAN), \cite{Aimr2018} proposes monocular depth estimation using synthetic data transformed from real-world RGB colour. As a multi-task approach, \cite{amir19} proposed temporally consistent depth prediction alongside semantic segmentation, which performed better than the single-task approach. Proposing an unsupervised depth estimation based on the ResNet-50 architecture, \cite{monodepth} uses the left image to generate depth for right-left images, and bilinear sampler and left-right disparity consistency loss to achieve significant improvement. 

Motivated by \cite{amir19}, we estimate depth using a monocular depth estimation component that includes two sub-models with shared weights, each model trained independently using shared weights but with each model being trained using either the \textit{normal} or \textit{foggy} datasets.
 
\subsection{Domain Adaptation}
\label{sec:style}
%\vspace{-.25cm}
In the current literature, domain adaptation has been used to transfer an image from its real domain to different domain (image-to-image translation) \cite{CycleGAN2017, pix2pix2016} allowing multiple uses of such images taken in complex environments or generated in different forms. 

The idea behind this approach is that the generated images from the source domain can be transformed to be similar to the ones in the target domain through capturing the style texture information of the input by utilising the Gram matrix. Work in \cite{amir_refer} shows that image style transfer (from the source domain to the target domain) is the process of minimising the differences between source and target distribution. Recent methods \cite{pix2pix2016, CycleGAN2017} use GAN \cite{gan} to learn the mapping from the source to the target images. Based on training over a large dataset for a specific image style,  \cite{CycleGAN2017} shows an efficient approach to transferring image style from one image into another.  

Another variation of domain adaptation has been performed within the same colour space of different domains (e.g. pixel-wise class labels for real-world and synthetic domains). In other words, the predictions derived from semantic segmentation components can be adapted by minimising the gap between them and the target ground truth \cite{tsai2018learning, amir19, wenzel2018modular}.

In this work, we employ the idea of \cite{CycleGAN2017} to map between \textit{normal} and \textit{foggy} weather conditions as a method to increase the degraded visibility level due to \textit{foggy} weather conditions. As a result, our model semantically segments a scene (taken in \textit{foggy} weather conditions) based on a synthetic \textit{normal} input (generated from \textit{foggy} scenes), which are considered as optimal inputs to the subsequent scene understanding process. As an additional step to achieving better segmentation performance, we use the technique proposed in \cite{wenzel2018modular, amir19} to jointly constrain depth estimation and semantic segmentation prediction close to the target domain (ground truth).

	\section{Proposed Approach}
\label{sec:method}

\begin{figure}[!tb]
	\centering
	\includegraphics[width=\linewidth]{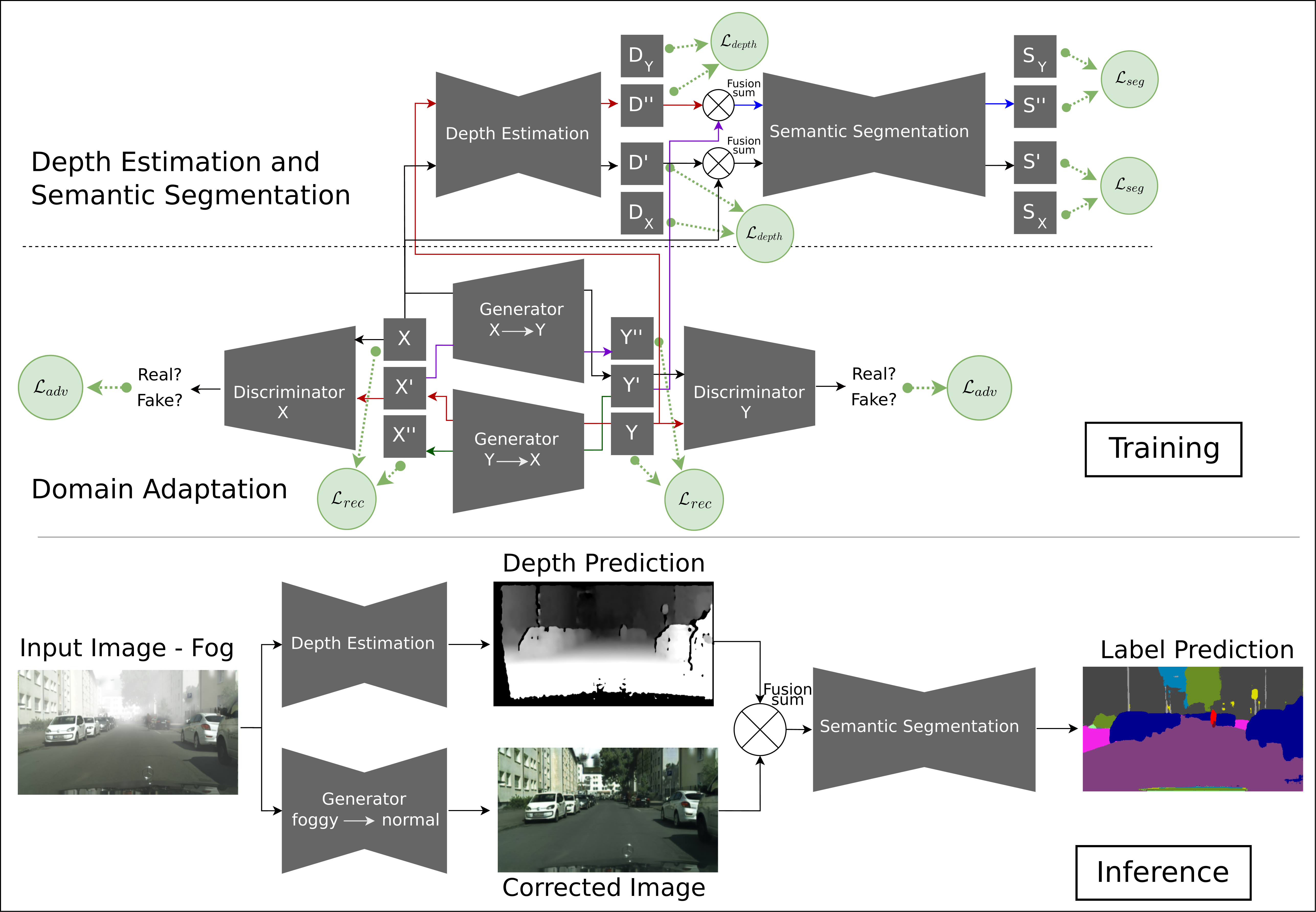} %overall_arch_toby width=14.5cm, height=10.0cm
	\caption{A conceptual overview of our scene understanding approach via domain adaptation using \cite{CycleGAN2017} (\textbf{Inference}) and the detailed outline of the entire pipeline (\textbf{Training}). Our overall model consists of two main components: domain adaptation using \cite{CycleGAN2017} and an encoder-decoder sub-module for semantic segmentation and depth estimation. Foggy scenes from (domain $X$) are transformed to fine scenes (domain $Y$) and vice versa, resulting in $Y', X'$ (the desired domains), and cyclically mapping them back to their original domains, producing $X'', Y''$; $D_{X}, D_{Y}$: ground truth depths and $D', D''$: depth predictions; $S_{X}, S_{Y}$ semantic labels and $S', S''$: semantic segmentation predictions.}
	 \vspace{-.55cm}  
	\label{fig:seg-depth-cycle}
\end{figure}

In simple terms, our main objective is to train an end-to-end network that semantically labels every pixel in a scene, and estimates the depth at each pixel from the monocular image, with both tasks occurring under foggy weather conditions. To achieve semantic segmentation under foggy weather conditions (the primary focus of our approach), we make use of knowledge adaptation  \cite{tsai2018learning} between models operating under \textit{normal} and \textit{foggy} weather conditions by simultaneously training two sub-models; each model is trained on one weather condition independently. Since scene visibility suffers due to foggy weather conditions, we make use of domain adaptation (Section \ref{sec:transfer}) as a method to increase the scene visibility level in the \textit{foggy} weather datasets, for semantic segmentation task. 

As an initial step towards improved semantic segmentation, monocular depth estimation is trained on both \textit{normal} and \textit{foggy} domains to produce depth maps for both domains. This step serves the semantic segmentation task by incorporating depth as a complementary information source with RGB colour \cite{ldfnet, fusenet}. In addition,  we consider using a multi-task approach as a feedback network \cite{amir19}, in which the output from a previous task serves as the input for the subsequent task, and the network recursively back propagates and updates its weights. Ultimately, the semantic segmentation is trained via two sub-models using \textit{normal} and synthetic \textit{normal} images (generated using the domain adaptation component in Section \ref{sec:transfer}).

In general, our approach consists of three sub-components: (1) Domain Adaptation (Section \ref{sec:transfer}), (2) Semantic Segmentation (Section \ref{sec:semeseg}), and (3) Monocular Depth Estimation (Section \ref{sec:depthEst})  (each functioning as an integrated unit). Our overall model produces three separate outputs: synthetic \textit{normal} images (generated from foggy weather condition), pixel-wise class labels, and depth. Figure \ref{fig:seg-depth-cycle} shows our proposed approach. In the remainder of this section, we discuss the details of the aforementioned three primary sub-components. 

\subsection{Domain Adaptation}
\label{sec:transfer}
Our goal of employing domain adaptation \cite{CycleGAN2017} (Figure \ref{fig:seg-depth-cycle} DA) in the context of semantic segmentation and monocular depth estimation is to increase the level of visibility under \textit{foggy} weather conditions via learning to map $\mathcal{D:} X \rightarrow Y$ from source domain $X$ (\textit{foggy} weather) to the target domain $Y$ (\textit{normal} weather) for which we assume such visibility corrected image is the optimal input to Semantic Segmentation (Section \ref{sec:semseg}). We GAN \cite{gan} with the cycle consistency method of \cite{CycleGAN2017} for mapping between \textit{foggy} and \textit{normal} weather conditions (Figure \ref{fig:seg-depth-cycle}). Two different generators $G_{X \rightarrow Y}$ (generating $Y'$), $G_{Y \rightarrow X}$ (generating $X'$) and two discriminators $D_{X}$ (to discriminate between $X$ and $X'$) , $G_{Y}$ (to discriminate between $Y$ and $Y'$) are used to perform the mapping function from the source and target domains. The loss for each generator $G$ with associated discriminator $D$ is as follows: 

% GAN loss
\begin{equation}\label{eq:adv_xy}
	\begin{aligned}
		\mathcal{L}_{adv}(X \rightarrow Y) = \min_{G_{Y \rightarrow X}} \max_{D_{Y}} \mathbb{E}_{y \sim \mathbb{P}_{d}(y)} [\log(D)_{(y)}]+ \\ \mathbb{E}_{x \sim \mathbb{P}_{d}(x)} [\log(1-D_{Y}(G_{X \rightarrow Y}(x)))],
	\end{aligned}
\end{equation}

\begin{equation}\label{eq:adv_yx}
\begin{aligned}
\mathcal{L}_{adv}(Y \rightarrow X) = \min_{G_{X \rightarrow Y}} \max_{D_{X}} \mathbb{E}_{y \sim \mathbb{P}_{d}(x)} [\log(D)_{(x)}]+ \\ \mathbb{E}_{x \sim \mathbb{P}_{d}(x)} [\log(1-D_{Y}(G_{Y \rightarrow X}(y))))],
\end{aligned}
\end{equation}where $\mathbb{P}_{d}$ is the data distribution, $X$ the source domain with samples $x$ and $Y$ the target domain with the samples $y$.

In addition to the adversarial loss $\mathcal{L}_{adv}$, a cycle-consistency loss $\mathcal{L}_{cyc}$ is used to map the transferred image ($Y'$) back to the source domain ($X$). The cycle-consistency loss is implemented as follows:

% cycle-consistency
\begin{equation}\label{eq:cycle_consistency}
    \begin{aligned}
    \mathcal{L}_{cyc} =  \lVert G_{Y \rightarrow X}(G_{X \rightarrow Y}(x)) - x \rVert_1 + \\ \lVert G_{X \rightarrow Y}(G_{Y \rightarrow X}(y)) - y \rVert_1
    \end{aligned}
\end{equation}

Subsequently, the joint loss function for the domain adaptation component is as follows:

% cycle-consistency
\begin{equation}\label{eq:domain-adapt}
\begin{aligned}
\mathcal{L}_{domain-adapt} = \mathcal{L}_{adv}(X \rightarrow Y) + \mathcal{L}_{adv}(Y \rightarrow X) + \mathcal{L}_{cyc}
\end{aligned}
\end{equation}

\subsection{Overall Segmentation and Depth Estimation Architecture}
As a subsequent task to domain adaptation in (Section \ref{sec:transfer}), semantic segmentation and depth estimation components are trained on two sets of scenes:- (1) real-world images $Y$ for \textit{normal} weather conditions (target domain) and (2) synthetic transferred images $Y'$ for \textit{foggy} weather conditions (source domain) (see Figure \ref{fig:seg-depth-cycle}, Inference). The transferred images represent the \textit{foggy} weather conditions $X$ which are mapped to the target domain \textit{normal} weather conditions $Y$ via $G_{X \rightarrow Y}(X)=Y'$. 

\begin{figure}[t!]
    \centering
    \includegraphics[width=\linewidth]{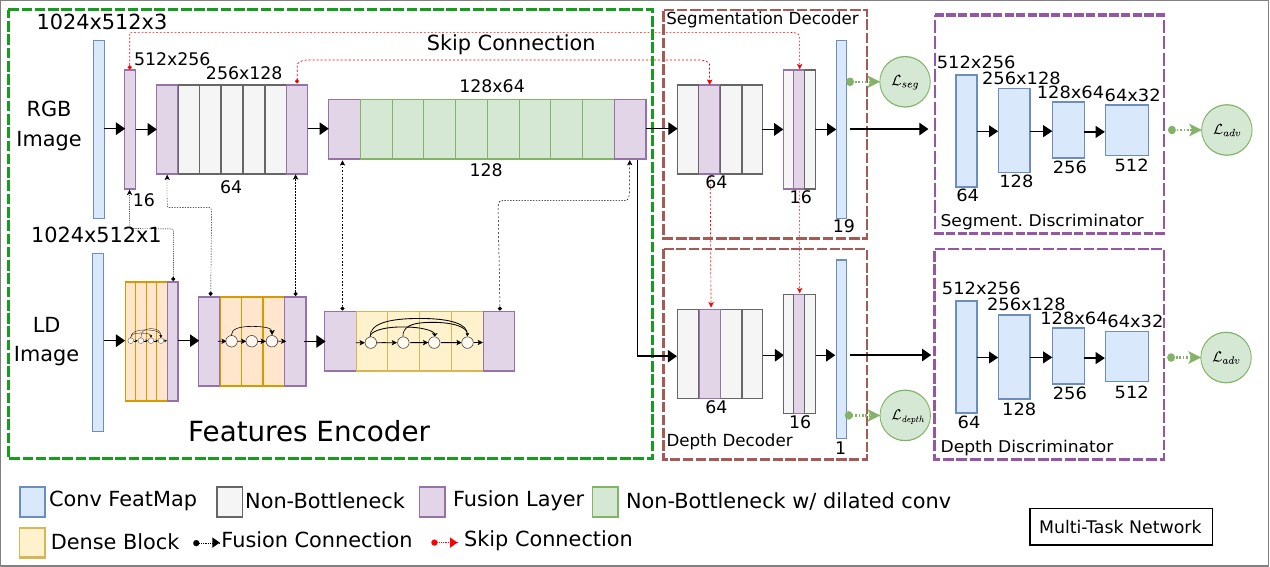}
    \caption{A detailed outline of the encoder-decoder architecture for semantic segmentation and depth estimation. The network consists of two sub-encoders taking two types of inputs: \textbf{RGB} and luminance \textbf{L} and/or depth \textbf{D} images (depending on the task); two decoders and two discriminators \cite{CycleGAN2017} for semantic segmentation and depth estimation.}
     \vspace{-.55cm}
    %\vspace{-20px}
    \label{fig:seg-net}
\end{figure}

%segmentation res
\begin{table*}[!t]
	\centering
	\resizebox{\linewidth}{!}{	
		\begin{tabular}{c|c|c|c|c|c|c|c|c|c}
			%\hline
			\multicolumn{4}{c|}{Methods} & \multicolumn{3}{c|}{Mean IoU} & \multicolumn{3}{c}{Complexity of the Network} \\ \hline \hline 
			{Models} & {Network Architecture} & Training & Fine-Tuning & \multicolumn{1}{c|}{Fog Zurich} & \multicolumn{1}{c|}{Fog Driving} & \multicolumn{1}{c|}{Fog Cityscapes} & \multicolumn{1}{c|}{Multi-Task} & \multicolumn{1}{c|}{Real-Time} & \multicolumn{1}{c}{Number of Parameters} \\ \hline \hline 
			CMDAda\cite{foggy19} & AdSegNet \cite{tsai2018learning} w/ DeepLab-v2 \cite{deeplab}      & C & \textemdash    & 25.0  & 29.7 & \textemdash & \textemdash & \textemdash & 44.0M  \\ \hline
			SFSU \cite{foggy18} & Dilated Conv. Net. (DCN) \cite{dcn}  & C & FC (498)       & 35.7 & 46.3 & \textemdash & \textemdash & \textemdash & 134M  \\ \hline 
			CMAda2+ \cite{foggy_old}& RefineNet \cite{refineNet17}  & C & FC (498)      & 43.4  & 49.9 & \textemdash & \textemdash & \textemdash & 118M \\ \hline 
			CMAda3+ \cite{foggy_old}  & RefineNet \cite{refineNet17}  & C & FC (498)       & 46.8 &  49.8 & \textemdash & \textemdash & \textemdash & 118M  \\ \hline
			
			Hanner \textit{et al.} \cite{foggy_pure} & RefineNet \cite{refineNet17}  & C & FS (24,500)         & 40.3 & 48.4 & \textemdash & \textemdash & \textemdash & 118M  \\ \hline
			Hanner \textit{et al.} \cite{foggy_pure} & RefineNet \cite{refineNet17}  & C & FS (498)         & 42.7 &  48.6 & \textemdash & \textemdash & \textemdash & 118M  \\ \hline
			Hanner \textit{et al.} \cite{foggy_pure} & RefineNet \cite{refineNet17}  & C & FC+FS (498)         & 41.4  & 50.7 & \textemdash & \textemdash & \textemdash & 118M  \\ \hline
			
			Hanner \textit{et al.} \cite{foggy_pure} & BiSeNet \cite{bisenet}  & C & FC (498)       & 25.0 &  30.3 & \textemdash & \textemdash & \textemdash & 50.8M \\ \hline
			Hanner \textit{et al.} \cite{foggy_pure} & BiSeNet \cite{bisenet}  & C & FS (24,500)    & 27.8 &  30.9 & \textemdash & \textemdash & \textemdash & 50.8M  \\ \hline
			Hanner \textit{et al.} \cite{foggy_pure} & BiSeNet \cite{bisenet}  & C & FS (498)       & 27.6 &  31.8 & \textemdash & \textemdash & \textemdash & 50.8M  \\ \hline
			Hanner \textit{et al.} \cite{foggy_pure} & RefineNet \cite{refineNet17}  & C & FC+FS (498)       & 35.2 &  30.9 & \textemdash & \textemdash & \textemdash & 118M  \\ \hline
			Ours w/o domain adaptation     & \textemdash & C & FC (498) &  13.9  & 17.6 & 59.4 & \textbf{\checkmark} & \textbf{\checkmark} & \textbf{4.8M} \\ \hline
			Ours w/ domain adaptation     & \textemdash & C & FC (498) &  26.1  & 31.6 & \textbf{60.3} & \textbf{\checkmark} & \textbf{\checkmark} & \textbf{16.2M} \\ \hline
			
	\end{tabular}}
	\vspace{3px}
	\caption{Quantitative comparison of semantic segmentation on Foggy Zurich \cite{foggy19}, Foggy Driving \cite{foggy18} and Foggy Cityscapes \cite{foggy18} datasets of our approach against state-of-the-art approaches. \textbf{C}: Cityscapes \cite{cityscapes}; \textbf{FC} Foggy Cityscapes \cite{foggy18}; \textbf{FS}: Foggy Synscapes \cite{foggy_pure}.} 
	\label{tab:main_res}
	\vspace{-.55cm}
\end{table*}

As seen in Figure \ref{fig:seg-net}, the overall segmentation and depth estimation architecture is designed with an auto-encoder which includes two distinct encoders: RGB encoder ($E_{RGB}$) and Luminance and/or Depth encoder ($E_{LD}$) to  extract features from RGB, Depth, and Luminance maps (Figure \ref{fig:seg-net}, Features Encoder). The two encoders are incorporated within the Features Encoder stage (Figure \ref{fig:seg-net}, Feature Encoder) to boost the performance of the multi-task model. $E_{RGB}$ and $E_{LD}$ are linked by fusing output layers from the corresponding blocks among $E_{RGB}$ and $E_{DL}$. The fusion connectivity is simply implemented by summing the two layers such that for inputs $x$ and $y$, the fused feature map is $E_{RGB}(x) + E_{DL}(y)$ or $E_{L}(y)$.  

Following the encoders, two decoders: the semantic segmentation decoder ($D_{Seg}$) (Figure \ref{fig:seg-net}, Segmentation Decoder) and the depth estimation decoder ($D_{Depth}$) (Figure \ref{fig:seg-net}, Depth Decoder) are designed to upsample the feature maps to the original input dimension for the two tasks of our model: pixel-wise segmentation with 19 class labels and depth images. Below we present in detail the encoders and decoders of the semantic segmentation and depth estimation components. 

\vspace{.2cm}
\noindent\textbf{{RGB encoder:}}
\label{sec:rgb_enc}
Designed to deal with a three-channel RGB input, the RGB encoder ($E_{RGB}$) (adopted from \cite{ldfnet}) comprises three downsampler blocks with convolutional and max pooling layers followed by batch normalization and \textit{ReLu()} activation function (\{\textit{16, 64, 128}\} respectively). Five non-bottleneck modules are implemented in the second downsampler block including the factorized convolutions (convolution kernel $n\times n$ factorized into $n\times 1$ and $1\times n$), each followed by batch normalization and \textit{ReLu()} with residual connections. With dilated and factorized convolutions in the third downsampler block, eight non-bottleneck modules with residual connections were utilised as a last component of $E_{RGB}$ to increase the RGB encoder efficiency.

\vspace{.1cm}
\noindent\textbf{{Luminance and/or Depth encoder:}}
\label{sec:ld_enc}  
Unlike the RGB encoder, the luminance and depth encoder ($E_{LD}$) (adopted from \cite{ldfnet}) deals with luminance and/or depth maps (concatenated as a two-channel input). We make use of a distinct encoder for luminance and depth to exploit better learning and representation from the depth and luminance maps that may not be possible when stacking depth with RGB colour as a four-channel image \cite{ldfnet, fusenet}. As a parallel function to $E_{RGB}$, $E_{LD}$ is designed using a dense connectivity technique for information flow enhancement from earlier to the final layers. Specifically, $E_{LD}$ consists of a downsampler (as in $E_{RGB}$) followed by three dense blocks; each has \{\textit{4, 3, 4}\} modules, respectively ($E_{LD}$ has the same number of channels as $E_{RGB}$). Each dense block is followed by a transition layer designed with $1\times1$ convolution layer and followed by $2\times2$ average pool layer. To achieve our goal in monocular depth prediction, we use only a luminance map in an encoder ($E_{L}$) which is identical to ($E_{LD}$) except than it takes the luminance channel only.

\vspace{.1cm}
\noindent\textbf{{Decoder:}}
\label{sec:seg_dec}
After fusing the last extracted feature maps from $E_{RGB}$ and either the $E_{LD}$ (for semantic segmentation) or $E_{L}$ (for monocular depth estimation), the depth decoder $(D_{Depth})$ and Semantic Segmentation decoder $(D_{Seg})$ perform upsampling upon the feature maps to the original resolution. This upsampling is implemented in three stages. In the first two stages \{\textit{64, 16}\}, convolutional transpose, batch normalisation and \textit{ReLu()} activation function, as well as two non-bottleneck modules, are employed. To this end, $D_{Seg}$ and $D_{Depth}$ perform the same process. As the last component in the $D_{Seg}$, a convolutional transpose layer maps the generated output from the previous layer to the 19 class labels we aim to predict. For monocular depth prediction, pyramid depth predictions are produced via  $D_{Depth}$ at two scales to gain consistent representation following \cite{monodepth}. Specifically, a convolutional transpose layer maps the predicted depth to match the original input dimension followed by a sigmoid activation function. Besides,  the previous stage (64 channels) are also mapped as in the final stage but at half the size of the original input dimension. 

\subsection{Semantic Segmentation}
\label{sec:semeseg}  

Our semantic segmentation model provides semantic predictions for two different scenes: (1) \textit{normal weather conditions}; and (2) \textit{foggy weather conditions}. To be more specific, two semantic segmentation sub-models with shared weights are trained on each weather conditions independently (\textit{normal} and \textit{foggy} weather conditions). We assume that sharing weights within the sub-models will allow transferring knowledge between \textit{normal} and \textit{foggy} domains and may contribute to improved segmentation in the later domain. As the scene visibility is very poor in the \textit{foggy} weather conditions, we use the corrected images $Y'$ (mapped from \textit{foggy} to \textit{normal} via domain adaptation \cite{CycleGAN2017} (Section \ref{sec:transfer})) as an alternative to \textit{foggy} scenes, assuming they are the optimal inputs to semantic segmentation.  

As a complementary information source, depth images are incorporated with RGB colour contributing to improved semantic segmentation performance \cite{fusenet, ldfnet}. As an initial step serving
semantic segmentation task, our model provides complementary depth images via monocular depth estimation (Section \ref{sec:depthEst}), which allows benefits from using depth with RGB colour to improve the performance of semantic segmentation. Serving the same goal, our semantic segmentation model uses the luminance input image, which is a translated grayscale image employed in \cite{ldfnet}. As a semantic segmentation loss function ($\mathcal{L}_{seg}$), cross-entropy is used. 

To force our semantic segmentation sub-model (\textit{foggy} weather conditions) to generate better segmentation labels close to performance under \textit{normal weather conditions}, we use an adversarial training approach \cite{gan} that is used in the literature \cite{Aimr2018, amir19, tsai2018learning, CycleGAN2017} to produce similar segmentation distributions in \textit{foggy} to \textit{normal} weather conditions. Specifically, we feed the predicted semantic labels from the segmentation sub-model (\textit{foggy} scenes) along with the corresponding ground truth labels into a discriminator ($D$) adapted from  \cite{CycleGAN2017} (Figure \ref{fig:seg-net}) to adapt output predictions by distinguishing predicted labels $G(x)=\tilde{y}$ from ground truth $y$. The adversarial loss ($\mathcal{L}_{adv}$) is used for our semantic segmentation and described in Eq. \ref{eq:adv_xy}. As an overall loss for the segmentation task, a joint segmentation loss defined as follows:

\begin{equation}\label{eq:joint_lossSeg}
    \mathcal{L}_{joint-seg} = \mathcal{L}_{seg} +
    \mathcal{L}_{adv}.
\end{equation}

%% depth res
\begin{table*}[t!]\centering
	\resizebox{\linewidth}{!}{
		\begin{tabular}{lccccccc} %{SSSSSSSS} {lccccccc}
			\hline \hline
			\multirow{2}[3]{*}{Method} & \multicolumn{4}{c}{Depth Error (lower, better)} & \multicolumn{3}{c}{Depth Accuracy(higher, better)} \\ 
			\cmidrule(lr){2-5} \cmidrule(lr){6-8}
			& {Abs. Rel.} & {Sq. Rel.}  & {RMSE}   & {RMSE log} & $\sigma$ $<$ $1.25$ & $\sigma$ $<$ $1.25^2$ & $\sigma$ $<$ $1.25^3$ \\ \hline \hline
			%\midrule
			Ours w/o domain adaptation & 0.238 & 0.543 &  1.994 & 0.277 & 0.656 & 0.884 & 0.983 \\
			Ours w/ domain adaptation & 0.238 & 0.733 &  2.130 & 0.280 & 0.654 & 0.892 & 0.980 \\
			\bottomrule
	\end{tabular}}
	\vspace{3px}
	\caption{Quantitative results of depth prediction over the \textit{refined Foggy Cityscapes} \cite{foggy18} with and without domain adaptation \cite{CycleGAN2017}.}
	\vspace{-.55cm}
	\label{tab:depth-measures}
\end{table*}  

\subsection{Monocular Depth Estimation}
\label{sec:depthEst}
Although monocular depth estimation is not the main objective of this paper, it has been used alongside semantic segmentation (our main objective) to improve the latter. Unlike when semantic segmentation performs individually, multi-modality allows us to gain deeper representation features in the overall model \cite{amir19} and perform inference in real-time \cite{cipolla2018multi}. In a similar vein to the earlier semantic segmentation component (Section \ref{sec:semeseg}), our model performs depth prediction via two sub-models over two scenes (\textit{normal} and \textit{foggy} weather conditions), each model on each weather condition. However, the depth estimation architecture deals only with RGB and luminance information as inputs. The loss function has been for depth estimation ($\mathcal{L}_{depth}$) is $\mathcal{L}_{1}$. We employ adversarial training to minimize the gap between the predicted depth on \textit{foggy} weather conditions against \textit{normal} weather conditions using a discriminator ($D$) takes predicted depth from the depth estimation sub-model (\textit{foggy} scenes) along with the corresponding ground truth to distinguish the predicted depth $G(x)=\tilde{y}$ from ground truth $y$. The adversarial loss ($\mathcal{L}_{adv}$) which described in Eq. \ref{eq:adv_xy} is used for depth estimation. As an overall loss for the depth estimation task, a joint depth loss is defined as follows: %Like the technique used in semantic segmentation,

\begin{equation}\label{eq:joint_lossDepth}
    \mathcal{L}_{joint-depth} = \mathcal{L}_{depth} +
    \mathcal{L}_{adv}.
\end{equation}

\subsection{Combined Loss} 
\label{sec:comb-loss}
Our combined loss function for the overall architecture with three sub-modules: domain adaptation, semantic segmentation, and depth estimation, is formulated in three steps. Firstly, adversarial loss for domain adaptation $\mathcal{L}_{adv}$ and cyclic-consistency loss ($\mathcal{L}_{cyc}$) functions are implemented. Secondly, we utilise the $\ell1$ loss for depth estimation with the adversarial loss for depth ($\mathcal{L}_{adv}$) on \textit{foggy} weather conditions. Finally, a cross-entropy loss is used as a semantic segmentation loss ($\mathcal{L}_{seg}$) as well as the adversarial loss for segmentation ($\mathcal{L}_{adv}$) on \textit{foggy} scenes.  The joint loss function on the overall architecture is thus as follows:    

% joint loss
\begin{equation}\label{eq:joint_loss}
    \begin{aligned}
        \mathcal{L} = \lambda\mathcal{L}_{domain-adapt} + \lambda\mathcal{L}_{joint-seg}+\lambda\mathcal{L}_{joint-depth},
    \end{aligned}
\end{equation}with $\lambda$ dynamically updated using the homoscedastic uncertainty technique
to weight and balance the losses \cite{cipolla2018multi}.

\subsection{Implementation Details}
\label{sec:implement}
Our implementation pipeline begins with the domain adaptation stage, followed by monocular depth estimation, then semantic segmentation stage. In domain adaptation, our goal is to generate corrected images from \textit{foggy} scenes (defogging process) to be used later in semantic segmentation. Therefore, we train two generators proposed in \cite{CycleGAN2017} on two domains (\textit{normal} and \textit{foggy}), each generator on each domain, to generate corrected images from the \textit{foggy} domain and close to the \textit{normal}. Subsequently, we trained the monocular depth estimation component via two sub-models using RGB and luminance inputs, each model on each weather condition independently, to produce depth used as a complementary information in semantic segmentation. Ultimately, we train the semantic segmentation component via two sub-models. One model is trained on \textit{normal} scenes using RGB, luminance and the generated depth map from the depth estimation stage. The other model is trained on the corrected images generated from \textit{foggy} scenes using domain adaptation, luminance, and the complementary depth information provided form depth estimation stage.  

\textit{Cityscapes} \cite{cityscapes} and the partially synthetic \textit{Foggy Cityscapes} \cite{foggy18} have been used as target and source domains, with $2,975$ training and $500$ testing image examples (at a resolution of $1024\times2048$). We applied data augmentation in training using random horizontal flip as well a down-sampled resolution of $128\times256$. In addition to \textit{Foggy Cityscapes}, real-world datasets: \textit{Foggy Driving} \cite{foggy18} and \textit{Foggy Zurich} \cite{foggy19} were used to evaluate our approach.  We implemented our approach in \textit{PyTorch} \cite{pytorch}. For optimization, we employed ADAM \cite{adam} with an initial learning rate of $1\times10^{-3}$ and momentum of $\beta_1 = 0.5, \beta_2 = 0.999$. Our model is optimized based on a joint loss discussed in Section \ref{sec:comb-loss}.

	\section{Experimental Results}
We evaluated the performance of our proposed approach on publicly available datasets:- \textit{Cityscapes dataset} \cite{cityscapes}, \textit{Foggy Cityscapes dataset} \cite{foggy18}, \textit{Foggy Driving} \cite{foggy18}, \textit{Foggy Zurich}, and \cite{foggy19} for semantic segmentation under foggy weather conditions. With and without using domain adaptation \cite{CycleGAN2017}, we assessed our approach using qualitative and quantitative comparisons against the state-of-the-art approaches. For semantic accuracy evaluation, we used the following evaluation measures: (1) class average accuracy, (2) global accuracy, and (3) mean intersection over union (mIoU). For the benchmark evaluation, we use the standard mIoU metric (Jaccard Index) which measures the percentage of mean intersections over union for predictions over all predicted classes. As an end-to-end pipeline, our model is trained to adapt \textit{foggy} to \textit{normal} weather conditions using \cite{CycleGAN2017} (Section \ref{sec:transfer}). Subsequently, depth estimation and semantic segmentation networks are trained. 
The detailed steps for the evaluation of our proposed architecture are as follows:

\renewcommand{\labelenumii}{\Roman{enumii}}
\begin{enumerate}
    \item \label{point1} We first train the domain adaptation component (Section \ref{sec:transfer}) on the \textit{Cityscapes} dataset (\textit{normal} weather) \cite{cityscapes} and Foggy Cityscapes (\textit{adverse} weather) \cite{foggy18} to map from \textit{adverse} scenes to \textit{normal} weather conditions.
    
    \item \label{point2} We train the depth estimation component (Section \ref{sec:depthEst}) on both the \textit{Cityscapes} dataset (\textit{normal} weather) \cite{cityscapes} and the Foggy Cityscapes datase (\textit{adverse} weather) \cite{foggy18} (two models for each with shared weights as set out in Section \ref{sec:depth}).
    
    \item \label{point3} Mirroring step \ref{point2}, we train the semantic segmentation component (Section \ref{sec:semeseg}), but using the corrected images from \textit{foggy} scenes generated from step \ref{point1} and incorporating the generated depth maps from step \ref{point2}.
    
    \item \label{point4} Models obtained from steps \ref{point1}, \ref{point2}, and \ref{point3} were fine-tuned using \textit{refined Cityscapes} \cite{foggy18} (a sub set that includes 498 training and 52 testing images examples with better quality).
    
    \item \label{point5} The fine-tuned models in step \ref{point4} were evaluated on both synthetic and real-world datasets including \textit{Foggy Zurich} \cite{foggy19} and \textit{Foggy Driving} \cite{foggy18}.
      
\end{enumerate}

In the rest of this section, we discuss the results of semantic segmentation (Section \ref{sec:seg-res}) and monocular depth estimation (Section \ref{sec:depth-res}).

\subsection{Semantic Segmentation}
\label{sec:seg-res}

\begin{figure}[t!]
    \centering
    \includegraphics[width=\linewidth]{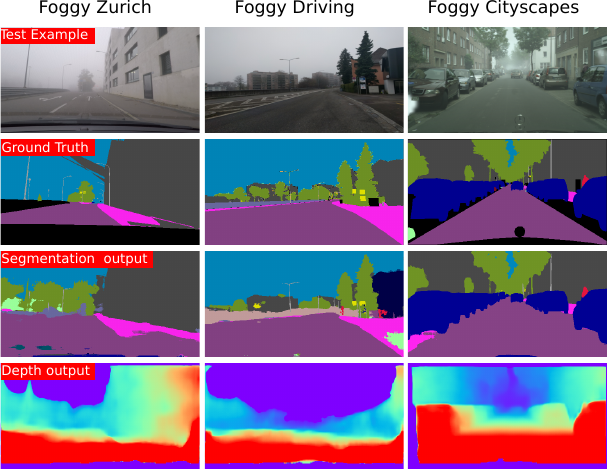}
    %\vspace{-.55cm}
    \caption{Segmentation and depth predictions on Foggy Zurich \cite{foggy18}, Foggy Driving \cite{foggy19} and Foggy Cityscapes \cite{foggy18} using our approach.}
    \label{fig:drivingandzurich-img}
    \vspace{-.55cm}
\end{figure}

We evaluated the performance of semantic segmentation on the following benchmark foggy weather conditions datasets: \textit{Foggy Driving} \cite{foggy18} and \textit{Foggy Zurich} \cite{foggy19}. This was a challenging
task as our model has not seen a single image from the aforementioned datasets. As an initial stage, we evaluated our model directly using \textit{foggy} scenes (with no \textit{domain adaptation}) from the aforementioned datasets. As seen in Table \ref{tab:main_res}, our model failed to obtain any favourable quantitative and qualitative results compared with no \textit{domain adaptation}. However, using \textit{domain adaptation}, our model clearly provides an improved performance of the mean intersection over union (mIoU) scores across all classes: from $13.9\%$ to \textbf{26.1}$\%$ on \textit{Foggy Zurich} \cite{foggy19} and from $17.8\%$ to \textbf{31.6}$\%$ of \textit{Foggy Driving} \cite{foggy18} (Table \ref{tab:main_res}). By contrast, we evaluate our model on a test set from (\textit{Foggy Cityscapes} \cite{foggy18}) having also trained on this dataset, which leads to improved segmentation: from $59.4\%$ to \textbf{60.3}$\%$ (Table \ref{tab:main_res}). Figure \ref{fig:drivingandzurich-img} shows qualitative results on \textit{Foggy Driving} \cite{foggy18}, \textit{Foggy Zurich} \cite{foggy19} and \textit{Foggy Cityscapes} \cite{foggy18} through different scenarios using our proposed approach. Overall, we consider that \textit{domain adaption}, as a method, influences the semantic segmentation performance under \textit{foggy} weather conditions. % (e.g. foggy weather conditions). 

As a comparison with the state-of-the-art semantic segmentation under foggy weather conditions, our approach with domain adaptation outperforms the work of \cite{foggy19, foggy_pure} on \textit{Foggy Zurich}. When tested on \textit{Foggy Driving} \cite{foggy18} our approach was able to surpass the work of \cite{foggy19}. In addition, our model outperforms the work of \cite{foggy_pure} with the three fine-tuned methods on: \textit{refined Foggy Cityscapes} \cite{cityscapes} ($498$) images, \textit{Foggy Synscapes} \cite{foggy_pure} ($24,000$) images,  and the combination of \textit{Foggy Cityscapes} \cite{cityscapes} and \textit{Foggy Synscapes} \cite{foggy_pure}.  However, our proposed approach remains competitive with the proposed approaches in \cite{foggy19, foggy18, foggy_pure, foggy_old}. Table \ref{tab:main_res} presents a comparison of our proposed approach against the-state-of-the-art foggy semantic segmentation.

Overall, we observe that our proposed approach provides a competitive performance against state-of-the-art techniques despite the complexity involve of using 
multi-task modality. In contrast, each component of the overall model has less
computational complexity. As clearly seen in Table \ref{tab:main_res}, the semantic segmentation component uses fewer parameters ($2.4$M) when compare with existing approaches, enabling the possibility of real-time performance. Another important aspect that underlines the superiority of our model is that all comparators use off-the-shelf complex segmentation networks such as RefineNet \cite{refineNet17}, DeepLab \cite{deeplab}, and Dilated Convolution Network \cite{dcn}, which constrained the practical application of their approaches in real-time performance.

\subsection{Monocular Depth Estimation}
\label{sec:depth-res}
Even though monocular depth estimation is not the primary focus,  we assess the efficacy of our model in monocular depth estimation using \textit{Cityscapes} \cite{cityscapes} and \textit{Foggy Cityscapes} which provide a disparity dataset labelled using Semi-Global Matching \cite{hirschmuller2007stereo}. Unlike the semantic segmentation network, the monocular depth component was not dependent on the \textit{domain adaptation} sub-model. In other words, we evaluate our model directly using \textit{Foggy} images, with and without \textit{domain adaptation} sub-model working alongside the depth estimation network (\textit{i.e.} no defogging or dehazing processing was used). Here, we assume that there is a similarity between fog and depth in terms of objects localization in which objects close to the camera are clearly visible. In the same vein, depth is used within the literature \cite{foggy18, foggy_pure} as a key input for fog simulation, whilst fog and noise lead to better depth estimation \cite{Aimr2018}. %Therefore, we took advantage from the aforementioned assumption to accelerate our model performance at both training and inference time as well as keeping the performance at the acceptable level.}
We evaluate our approach on monocular depth estimation quantitatively and qualitatively using two methods. Firstly, we use a single model to perform the following three tasks: (1) domain adaptation, (2) semantic segmentation, and (3) monocular depth estimation. Secondly, we use the same model but without the domain adaptation component. Measurement metrics are based on \cite{eigen2014depth}.  As seen in Table \ref{tab:depth-measures}, our approach provides monocular depth estimation results that are close to each other using the two aforementioned methods.

	\section{Conclusion}
We propose a novel multi-task approach for automotive semantic segmentation and depth estimation under \textit{foggy} weather conditions. Our approach is designed via multi-modality to produce optimal performance through using domain adaptation (GAN-based) \cite{CycleGAN2017} to correct images with poor visibility taken in \textit{foggy} weather conditions. By using synthetic and real-world datasets, depth estimation and semantic segmentation components are trained with a unified framework providing promising results. With dense-connectivity, skip-connections, and fusion-based techniques, we propose a competitive encoder-decoder for semantic segmentation and depth estimation were proposed. Our overall approach is characterized by a complexity that allows multi-task learning. In addition, each component was designed with a lightweight architecture allowing real-time performance. Using extensive experimentation, we show the performance of our approach achieves significant results over the state-of-the-art semantic segmentation under adverse weather condition \cite{foggy18, foggy19, foggy_pure} as well as providing extra tasks (\textit{i.e.,} monocular depth estimation). %.This paper proposes a novel end-to-end automotive semantic segmentation within foggy scene understanding. Using a unified model, we make use of domain adaptation (GAN-based) \cite{CycleGAN2017} to adapt a scene taken in \textit{foggy} weather conditions to \textit{normal} thus increasing the scene visibility. Subsequently, the adapted images are fed to an effective semantic segmentation model for training. For real-time performance, our segmentation network is based on light-weight architecture that includes features fusion, dense connectivity and skip connections, making the approach real-time ($20$ -- $42$ fps with and without \textit{domain adaptation} respectively). As a result, the performance of our approach has progressively improved and achieved significant performance over the state-of-the-art semantic segmentation under foggy weather conditions \cite{foggy18, foggy19, foggy_old}.

%We propose a novel multi-task approach for automotive semantic segmentation and depth estimation under \textit{foggy} weather conditions. Our approach is designed via multi-modality to produce optimal performance through using domain adaptation (GAN-based) \cite{CycleGAN2017} to correct images with poor visibility taken in \textit{foggy} weather conditions. By using synthetic and real-world datasets, depth estimation and semantic segmentation components are trained with a unified framework providing promising results. With dense-connectivity, skip-connections, and fusion-based techniques, we propose a competitive encoder-decoder for semantic segmentation and depth estimation were proposed. Our overall approach is characterized by a complexity that allows multi-task learning. In addition, each component was designed with a lightweight architecture allowing real-time performance. Using extensive experimentation, we show the performance of our approach achieves significant results over the state-of-the-art semantic segmentation under adverse weather condition \cite{foggy18, foggy19, foggy_pure} as well as providing extra tasks (\textit{i.e.,} monocular depth estimation).
	{\small
\bibliographystyle{lib/IEEEtranS}
\bibliography{ref/egbib}
} % references

% that's all folks
\end{document}